\documentclass[10pt,twocolumn,letterpaper]{article}

\usepackage{cvpr}              

\newcommand{\anonymousHospital}{Sankara Eye Hospital}
\newcommand{\msicsDataset}{Sankara-MSICS}

\usepackage{multirow}
\definecolor{cvprblue}{rgb}{0.21,0.49,0.74}
\usepackage[pagebackref,breaklinks,colorlinks,allcolors=cvprblue]{hyperref}
\usepackage{xcolor}
\usepackage{booktabs} 
\usepackage{colortbl} 
\usepackage{xcolor}   
\usepackage{adjustbox} 
\def\paperID{16642} 
\def\confName{CVPR}
\def\confYear{2025}

\title{Phase-Informed Tool Segmentation for Manual Small-Incision Cataract Surgery}

\author{
Bhuvan Sachdeva$^{1, 2}$\thanks{Equal contribution.}, 
Naren Akash$^{1}$\footnotemark[1], 
Tajamul Ashraf$^{1}$,
\and
Simon Mueller$^{3}$, 
Thomas Schultz$^{3}$, 
Maximilian W. M. Wintergerst$^{3}$, 
\and
Niharika Singri Prasad$^{2}$, 
Kaushik Murali$^{2}$\thanks{Corresponding authors: kaushik@sankaraeye.com and \\ mohja@microsoft.com.}, 
Mohit Jain$^{1}$\footnotemark[2]\\
\\
\small $^{1}$Microsoft Research, Bengaluru, India\\
\small $^{2}$Sankara Eye Hospital, Bengaluru, India\\
\small $^{3}$University of Bonn, Bonn, Germany
}

\begin{document}
\maketitle
\begin{abstract}
\textit{Cataract surgery is the most common surgical procedure globally, with a disproportionately higher burden in developing countries. While automated surgical video analysis has been explored in general surgery, its application to ophthalmic procedures remains limited. Existing works primarily focus on Phaco cataract surgery, an expensive technique not accessible in regions where cataract treatment is most needed. In contrast, Manual Small-Incision Cataract Surgery (MSICS) is the preferred low-cost, faster alternative in high-volume settings and for challenging cases. However, no dataset exists for MSICS. To address this gap, we introduce \msicsDataset{}, the first comprehensive dataset containing 53 surgical videos annotated for 18 surgical phases and 3,527 frames with 13 surgical tools at the pixel level. We benchmark this dataset on state-of-the-art models and present ToolSeg, a novel framework that enhances tool segmentation by introducing a phase-conditional decoder and a simple yet effective semi-supervised setup leveraging pseudo-labels from foundation models. Our approach significantly improves segmentation performance, achieving a $23.77\%$ to $38.10\%$ increase in mean Dice scores, with a notable boost for tools that are less prevalent and small. Furthermore, we demonstrate that ToolSeg generalizes to other surgical settings, showcasing its effectiveness on the CaDIS dataset.} 
\end{abstract}
\section{Introduction}
\label{sec:intro}

\begin{figure}[!t]
    \centering
    \includegraphics[width=1.0\linewidth]{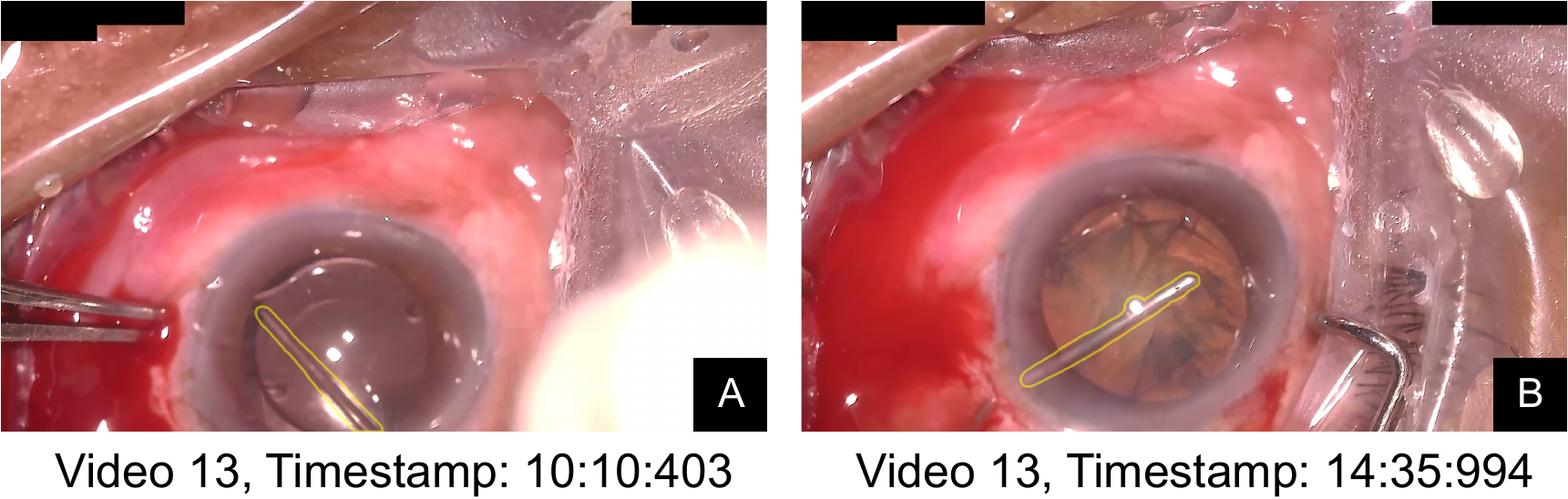}
    \caption{
        Although the surgical tools in Images A and B appear similar, their surgical phases differ. Image A shows a Hydrodissection Cannula (from the Hydroprocedure phase), while Image B shows a Dialer (from the OVD, IOL Insertion phase). ToolSeg, our proposed method, leverages surgical phase information to accurately classify and segment instruments.}
    \label{fig:teaser}
    
\end{figure}

Cataract is the leading cause of preventable blindness worldwide, with surgical intervention being the standard treatment. Over 26 million individuals undergo cataract surgery annually~\cite{burton2021lancet,cicinelli2023cataracts}, making it one of the most common surgical procedures.
Established techniques include Phacoemulsification (Phaco), Manual Small Incision Cataract Surgery (MSICS), and Femto-Laser-Assisted Cataract Surgery (FLACS). Unlike laparoscopic surgery, where most computer vision methods have been developed~\cite{rodrigues2022surgical}, cataract surgery presents unique challenges. It involves delicate micro-instruments in a highly reflective ocular environment~\cite{meek2015corneal}, resulting in specular reflections that distort the visual appearance of surgical tools~\cite{curlin2020current,boldrey1984retinal}. Additionally, the small instruments cause significant foreground imbalance (e.g., Figure~\ref{fig:teaser}), and the transparent ocular tissues, combined with microscope use, create complex optical conditions, thus complicating image analysis.

There is an increasing need to expand surgical capacity, safety, and efficiency~\cite{dobson2020trauma,varghese2024artificial}. Although AI-driven tools for surgery are still in their nascent stages, they hold the potential to significantly improve every aspect of surgical care. In particular, automatic (real-time) surgical video analysis can advance surgeon skill development, improve training with targeted feedback, ensure quality control, and detect surgical anomalies~\cite{tollefson2024defining}. Achieving this requires robust temporal and spatial video analysis for automated surgical workflow and scene understanding, which depends on accurately identifying surgical phases and precisely segmenting surgical tools.

Deep learning-based computer vision algorithms have been extensively developed for analyzing cataract surgeries, but their application has largely been limited to Phaco procedures~\cite{muller2024artificial}. While Phaco is the preferred technique in developed countries, the burden of cataract blindness remains disproportionately high in low- and middle-income countries (LMICs)~\cite{fang2022global}. Despite Phaco's efficacy, its reliance on expensive technology and infrastructure limits accessibility in these regions. In contrast, MSICS is the most commonly performed procedure in resource-limited settings due to its lower cost, faster execution, and reduced reliance on technology, making it well-suited for high-volume surgeries~\cite{Spencer2006, yorston2005high}. MSICS is also the preferred method for challenging cases, such as brunescent hard, hypermature, and intumescent cataracts, as well as for patients with loose zonules~\cite{Alio2022-oc}.
Despite its prevalence and advantages, MSICS has been largely neglected in the development of datasets for automated ophthalmic surgical video analysis. To date, no publicly available dataset exists specifically for MSICS~\cite{muller2024artificial}.

In this paper, we introduce \textit{\msicsDataset{}}, the first large-scale dataset on manual small incision cataract surgery. This dataset includes 3,527 frames from 53 \textit{in vivo} human cataract surgery videos, annotated with pixel-level labels for 13 surgical tools and corresponding phases across 18 surgical stages.

We benchmark \textit{\msicsDataset{}} using state-of-the-art methods for surgical tool segmentation, achieving $52.03\%$ Dice score (DSC),
which highlights the need for specialized approaches.
Our analysis of the \msicsDataset{} dataset reveal a strong correlation between surgical phases and tool presence.
Building on this insight, we propose \textit{ToolSeg}, a novel framework that leverages surgical phase information as a prior for tool segmentation. 
\textit{ToolSeg} outperforms existing methods by $24.46\%$ DSC, with particularly notable improvements in classifying and segmenting less prevalent tools.
We also leverage Meta's SAM 2 model~\cite{ravi2024sam} to expand our dataset by propagating labels forward and backward in time across videos from a single annotated frame, substantially increasing labeled frames (to 24,405 frames) without incurring high annotation costs and resulting in a further improvement of $8.02\%$ DSC.
Additionally, we validated our \textit{ToolSeg} and SAM 2 based approach to the publicly available CaDIS dataset~\cite{grammatikopoulou2021cadis} for Phaco surgeries demonstrating the generalizability of our proposed approaches.

The main contributions of this paper are:
\begin{itemize}[noitemsep]
    \item We introduce the \textbf{\msicsDataset{}}, the first comprehensive dataset tailored for Manual Small-Incision Cataract Surgery, comprising of 3,527 frames from 53 surgical videos with ground truth annotations for 18 surgical phases and 13 surgical tools. This dataset fills a critical gap in cataract surgery analysis, enabling research in low-cost, high-volume surgical contexts.
    \item We propose \textbf{ToolSeg}, a novel framework that incorporates surgical phase information as a prior through a phase-conditional decoder. ToolSeg outperforms previous methods with an average IoU improvement of $31.27\%$ and DSC improvement of $24.46\%$ on \msicsDataset{}, thus demonstrating the value of using surgical phase information to enhance instrument segmentation.
    \item We develop a simple yet effective semi-supervised learning setup leveraging pseudo-labels from foundation models (SAM 2). This approach generates high-quality labels from the manually labeled frames using mask propagation and iterative prompting, resulting in a substantial boost of $31.27\%$ IoU and $24.47\%$ DSC on \msicsDataset{}.
\end{itemize}

\section{Related Work}
\label{sec:rw}

\subsection{Surgical Tool Datasets}

The American College of Surgeons recognizes fourteen surgical specialties~\cite{ban2017american}, 
each with unique procedures and requiring specific sets of surgical tools, including specialized instruments, implants, and screws tailored to particular anatomical regions. While there has been significant progress in the field of surgical image analysis, most existing datasets focus on robotic and laparoscopic surgeries, particularly in gastroenterology. For instance, the EndoVis 2015 challenge dataset provides 2D images from four colorectal surgeries with three tool classes~\cite{bodenstedt2018comparative}. M2CAI~\cite{twinanda2016endonet,maqbool2020m2caiseg} offers 15 cholecystectomy (gall bladder removal) videos with binary annotations for seven tools. 
The ROBUST-MIS dataset~\cite{ross2020robust} includes over 10,000 labeled frames across two classes for procedures like rectal resection and proctocolectomy.
For cataract surgery, however, public datasets are limited and focus exclusively on Phaco~\cite{mahmud2015proposed}. Cataract surgery presents unique challenges, as it involves delicate micro-instruments in a highly reflective ocular environment~\cite{al2019cataracts}. Specular reflections distort the visual appearance of these tools, and partial tool insertion into the eye cavity can render parts nearly invisible. This setting, combined with transparent ocular tissues and microscope use for surgery, results in significant foreground-background imbalance and complex optical conditions, thus further complicating image analysis~\cite{al2019cataracts}. 
Existing Phaco datasets include InSegCat~\cite{schoeffmann2018cataract,fox2020pixel} with annotations for 11 tools classes in 843 frames, Cataracts-1K~\cite{ghamsarian2024cataract} providing nine tool classes in 2,256 frames from 30 videos, and CaDIS~\cite{grammatikopoulou2021cadis} offering 29 tool classes in 4,671 frames.
In contrast, MSICS is the preferred procedure in resource-limited settings due to its affordability, speed, lower expertise requirement, and reduced dependence on technology~\cite{singh2017review}, making it suitable for high-volume surgeries. MSICS is also favored for challenging cases like hard or hypermature cataracts and patients with weak zonules~\cite{Alio2022-oc}. Unlike Phaco, which requires a 2-3 mm incision, MSICS involves a larger 6-8 mm incision, allowing for easier maneuvering of tools. The surgical phases in MSICS are distinct, with steps such as Nucleus Delivery, Nucleus Prolapse, and Peritomy not performed in Phaco, which instead includes Trenching, Nucleus Emulsification, and Irrigation/Aspiration~\cite{Spencer2006}. MSICS also utilizes specialized tools such as the Vectis, Dialer, Conjunctival Scissors, Simcoe Cannula, Cautery, and Crescent Blade, which are not used in Phaco; conversely, Phaco uses tools like Phaco Probe, Irrigation/Aspiration Probe, and Lens Injector~\cite{grammatikopoulou2021cadis}.
Despite its widespread adoption, no publicly available dataset exists for MSICS~\cite{muller2024artificial}, leaving a critical gap in cataract surgery research. Our work is the first to address this need by introducing a comprehensive, phase- and tool-annotated dataset for MSICS, essential for advancing surgical tool segmentation in resource-limited settings.

\subsection{Surgery Tool Segmentation}
Early works on surgical tool segmentation~\cite{ronneberger2015u, iglovikov2018ternausnet} primarily focused on \textit{semantic segmentation}, aiming to classify each pixel independently. Building on the U-Net~\cite{ronneberger2015u} encoder-decoder architecture, subsequent methods introduced modifications to enhance segmentation performance. For instance, TernausNet~\cite{shvets2018automatic} incorporated a VGG backbone for improved feature extraction.
Attention mechanisms were explored in RASNet~\cite{ni2019rasnet} and RauNet~\cite{ni2019raunet} to focus on target regions, while AP-MTL \cite{islam2020ap} refined this with global attention dynamic pruning.
PAANet~\cite{ni2020pyramid} and SurgiNet~\cite{ni2022surginet} incorporated multi-scale pyramid attentive features to improve segmentation. 

Despite these advancements, these methods encountered challenges in accurately localizing and differentiating surgical tools from the background. Attempts have been made to incorporate prior information to improve the segmentation performance. Boundary-based approaches, such as ToolNet~\cite{garcia2017toolnet}, captured spatial extent of surgical tools, while saliency maps in MTL~\cite{islam2019learning} and ST-MTL \cite{islam2021st} helped focus attention on instrument-relevant regions. 
To exploit temporal dependencies from the sequential data, methods like MF-TAFNet~\cite{jin2019incorporating} incorporated flow-based temporal priors. While these approaches yielded improvements, semantic segmentation suffers from inherent limitations such as local inconsistency---where a single tool might not be segmented as a complete, integral object---and semantic inconsistency---where a single instrument can be assigned multiple instrument types---leading to spatial class inconsistency within objects.

\textit{Instance segmentation} methods were developed as an alternative paradigm to semantic segmentation for surgical tool segmentation. It involves predicting a set of masks and then associating each mask with a class label, inherently reducing spatial class inconsistency. Building upon the success of Mask R-CNN~\cite{he2017mask}, several works adapted this framework to the surgical domain. \citet{kong2021accurate} fine-tuned Mask R-CNN model optimized with anchor optimization and improved region proposal network. ISINet~\cite{gonzalez2020isinet} builds upon Mask R-CNN by incorporating a relabeling strategy that leverages temporal redundancy in prediction to correct mislabelled surgical tools. 
Transformer-based methods gained prominence due to their ability to capture long-range dependencies. TraSeTR~\cite{zhao2022trasetr} introduced a transformer-based track-to-segment framework that incorporates tracking cues for improved instance segmentation. MATIS~\cite{ayobi2023matis} leveraged the transformer architecture by adapting Mask2Former \cite{cheng2021per,cheng2022masked} for surgical tool segmentation and incorporating temporal consistency to enhance performance.
In this work, we benchmarked our model using approaches like U-Net, ISINet, and MATIS.
\section{The \msicsDataset{} Dataset}
\label{sec:cataract-sics-dataset}

\begin{figure}[t]
    \centering
    \includegraphics[width=1.0\linewidth]{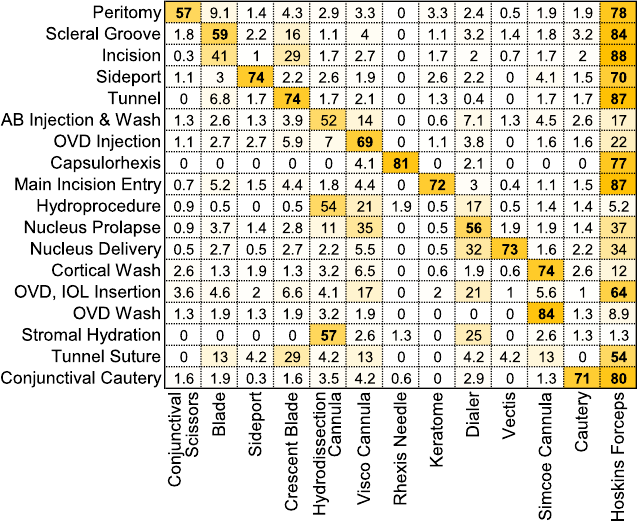}
    \caption{Surgical phase and surgical tool co-occurence.}
    \label{fig:phase_tool_correl}
\end{figure}

The \msicsDataset{} dataset consists of 53 cataract surgery videos recorded at \anonymousHospital{}, in \textit{AnonymizedLocation}, from October 2023 to October 2024. Each video, averaging $15 \text{ min } 39 \text{ s} \pm 7 \text{ min } 38 \text{ s}$, was recorded at 30 fps with a resolution of 1920 x 1080 using a microscope-mounted video camera. Note: The surgeons performed the surgery using the same microscope view as recorded in the videos. 
To optimize computation, all frames were downscaled by a factor of four to a resolution of 480 x 270.

Two resident ophthalmologists at \anonymousHospital{} discussed and defined 18 surgical phases of MSICS surgery, and annotated all videos with start and stop timestamps for each phase.
Frames were uniformly extracted across phases, but additional frames featuring less frequently used surgical tools were added to address class imbalance.
The extracted frames were then segmented and labeled by resident ophthalmologists.

For annotation, we developed a web-based portal based on Meta's Segment Anything Model (SAM)~\cite{kirillov2023segment}.
Annotators segmented each tool by prompting SAM with positive and negative clicks and selecting the tool class from a drop-down list of 13 tools, minimizing manual effort.
Each frame was provided with its corresponding surgical phase to help annotators accurately identify tools.
Annotations were conducted in sessions with 6-8 annotators working in 2-hour blocks, each annotating 120-150 frames. For quality control, all annotators were assigned a common subset of 30 frames, initially annotated by a senior ophthalmologist. This set was programmatically and manually compared to the ground truth annotations, resulting in the rejection of annotations from two ophthalmologists across sessions due to quality issues.
As the segmentation was SAM-based, we implemented a denoising pipeline using morphological opening and closing operations to remove artifacts and smooth the annotated masks.

The final dataset consists of 3,527 frames annotated for 13 surgical tools (at pixel-level) and 18 surgical phases, as detailed in Table~\ref{tab:classwise-comparison}. As expected, the dataset follows a long-tail distribution, with three tools having fewer than 200 instances.
\section{ToolSeg: Our Proposed Method}

\subsection{Phase Tool Correlation}
Analysis of the \msicsDataset{} dataset reveals a strong correlation between surgical phases and tool presence (Figure~\ref{fig:phase_tool_correl}).
For instance, tool \textbf{Vectis} was exclusively used in phase \textbf{Nucleus Delivery}, or tool \textbf{Cautery} was most commonly used in phase \textbf{Conjunctival Cautery}.
On the other hand, instruments such as \textbf{Hoskins Forceps} and \textbf{Crescent Blade} are used in multiple surgical phases.
Based on these insights, we propose an approach that leverages surgical phase information as a prior for tool segmentation.
\begin{figure}[t]
    \centering
    \includegraphics[width=1.0\linewidth]{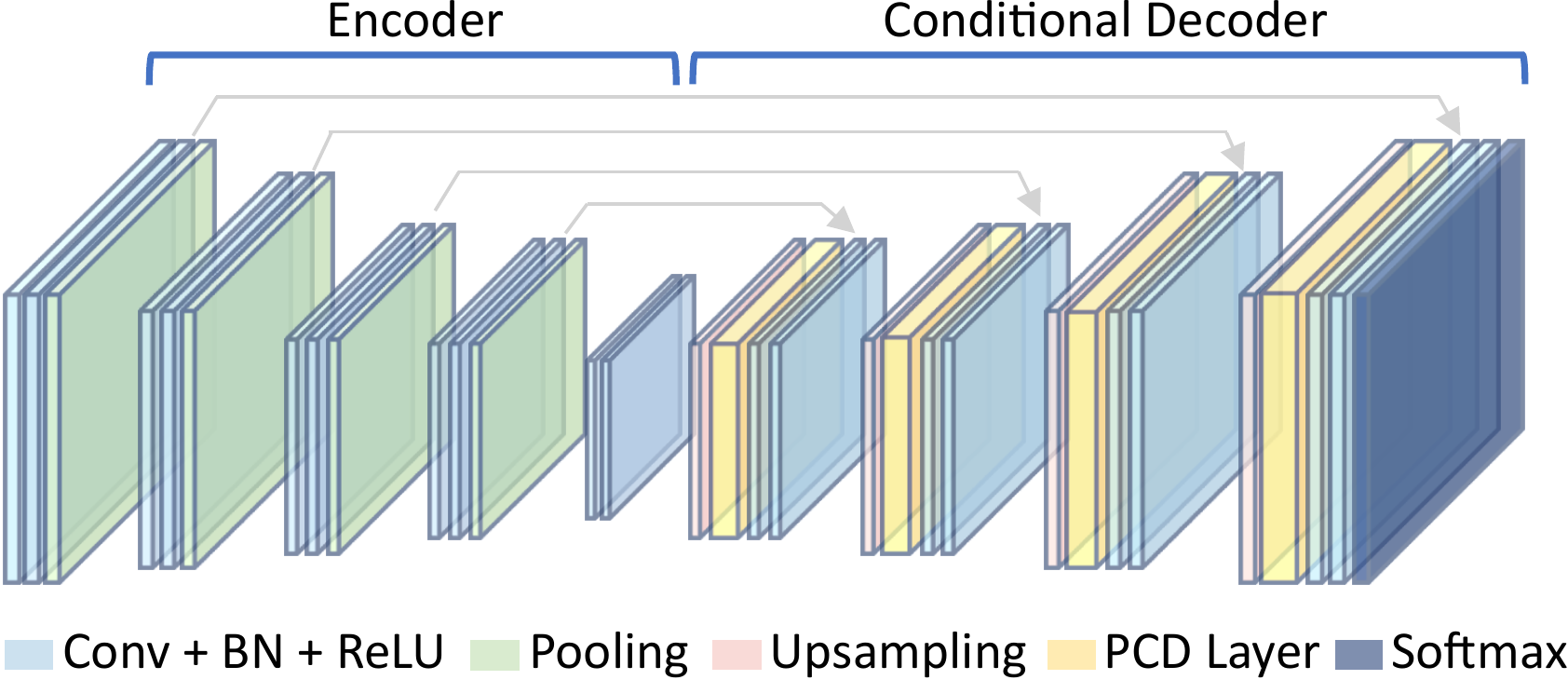}
    \caption{An overview of the ToolSeg network architecture.}
    \label{fig:toolseg_encdec}
    
\end{figure}

\subsection{Preliminary: Surgery Phase Recognition}
To leverage surgical phase information for tool segmentation, we first employ the Multi-Stage Temporal Convolutional Network (MS-TCN++)~\cite{li2020ms} architecture to predict the phase of each frame.
MS-TCN++ has achieved state-of-the-art performance on cataract surgery datasets~\cite{li2020ms}. It was originally proposed for temporal action segmentation, offering efficient and real-time phase recognition from long, untrimmed surgical videos. This architecture processes the video at full temporal resolution, enabling smooth and consistent predictions through its multi-stage design.
The initial stage with dual dilated layers generates a preliminary phase prediction, while subsequent stages with dilated residual layers iteratively refine this prediction. Given an input video, the model outputs a phase label for each frame.

\subsection{Phase-Conditioned Segmentation Network}

\begin{figure}[t]
    \centering
    \includegraphics[width=1.0\linewidth]{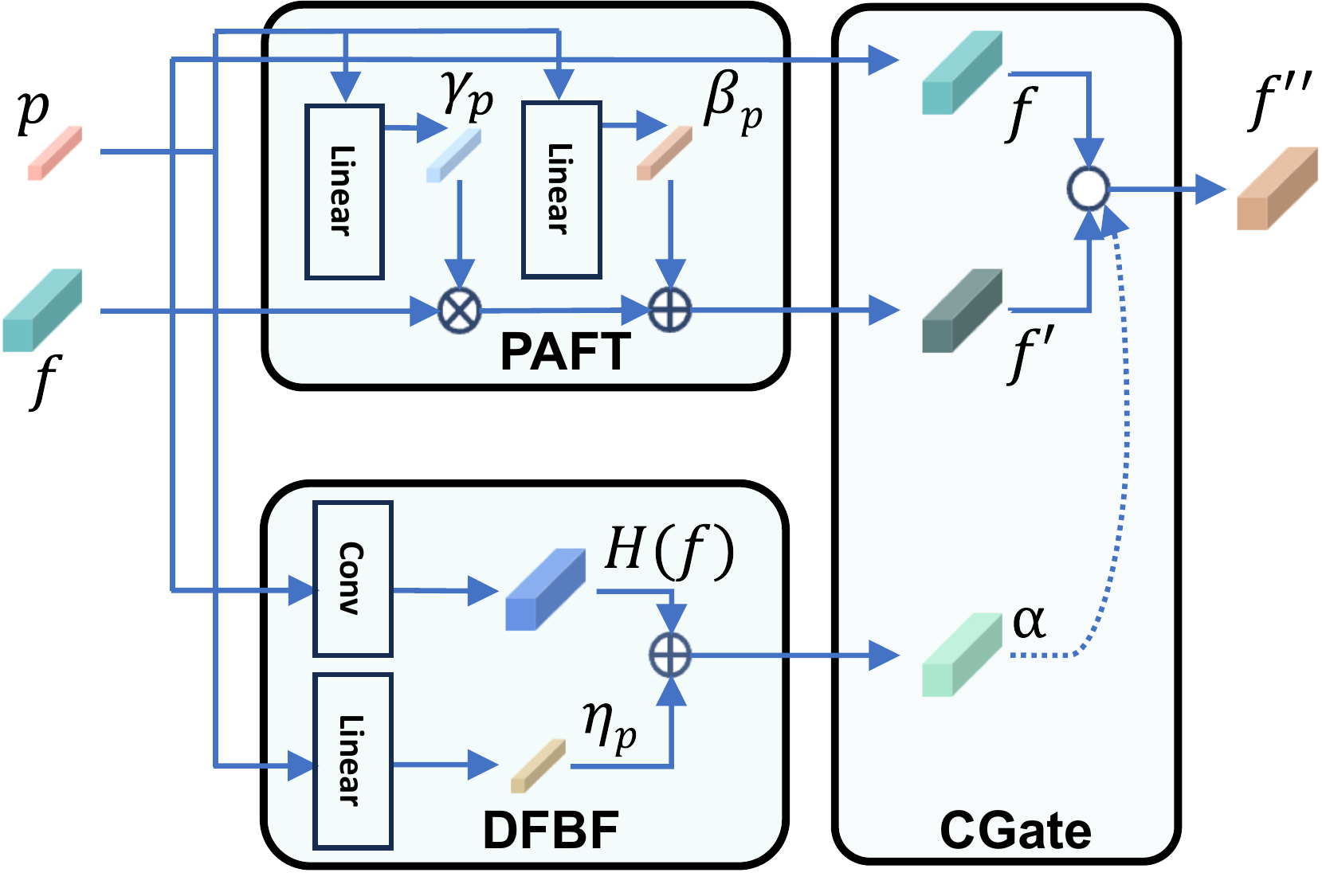}
    \caption{Key components of Phase-informed Conditional Decoder layer: PAFT, DFBF, and CGate.}
    \label{fig:toolseg_encdeccomp}
\end{figure}

We propose \textit{ToolSeg}, an encoder-decoder architecture for surgical tool segmentation, in which the decoder is conditioned on the surgical phase to improve segmentation accuracy. To achieve this, we introduce the Phase-informed Conditional Decoder (PCD) layer, which is added at each decoder level and consists of three key components: Phase-aware Affine Feature Transform (PAFT), Dynamic Feature Blending Factor (DFBF), and Context-Aware Adaptive Gating (CGate). PAFT modulates feature maps channel-wise, while DFBF applies spatial modulation. CGate combines these modulations to improve segmentation. We propose two variants of phase-specific conditioning in the PCD layer: \textit{Basic Conditioning} (\textbf{PCD-Basic}), where only PAFT is applied, and \textit{Gated Conditioning} (\textbf{PCD-Gated}), where PAFT is combined with DFBF and CGate to further enhance feature modulation and segmentation accuracy. A schematic of the framework is shown in Figure~\ref{fig:toolseg_encdec} and \ref{fig:toolseg_encdeccomp}. \\

\noindent \textbf{Phase-aware Affine Feature Transform.} 
To leverage phase-specific information for tool localization, the PAFT module conditions the segmentation network on the predicted surgical phase by learning phase-specific channel-wise shift and scale embeddings. For each phase $p$, a pair of learnable embeddings---$\gamma_p$ (for shift) and $\beta_p$ (for scale)---captures phase-related priors, such as phase-tool correlation and tool co-occurrence. These embeddings are applied to each input feature map $f$ as:
\[ f^\prime = \gamma_p \odot f + \beta_p \]
This enables the network to adaptively adjust its feature maps \textit{channel-wise} based on the current phase, improving tool segmentation by emphasizing phase-relevant features.\\

\begin{figure*}[t!]
    \centering
    \includegraphics[width=1.0\linewidth]{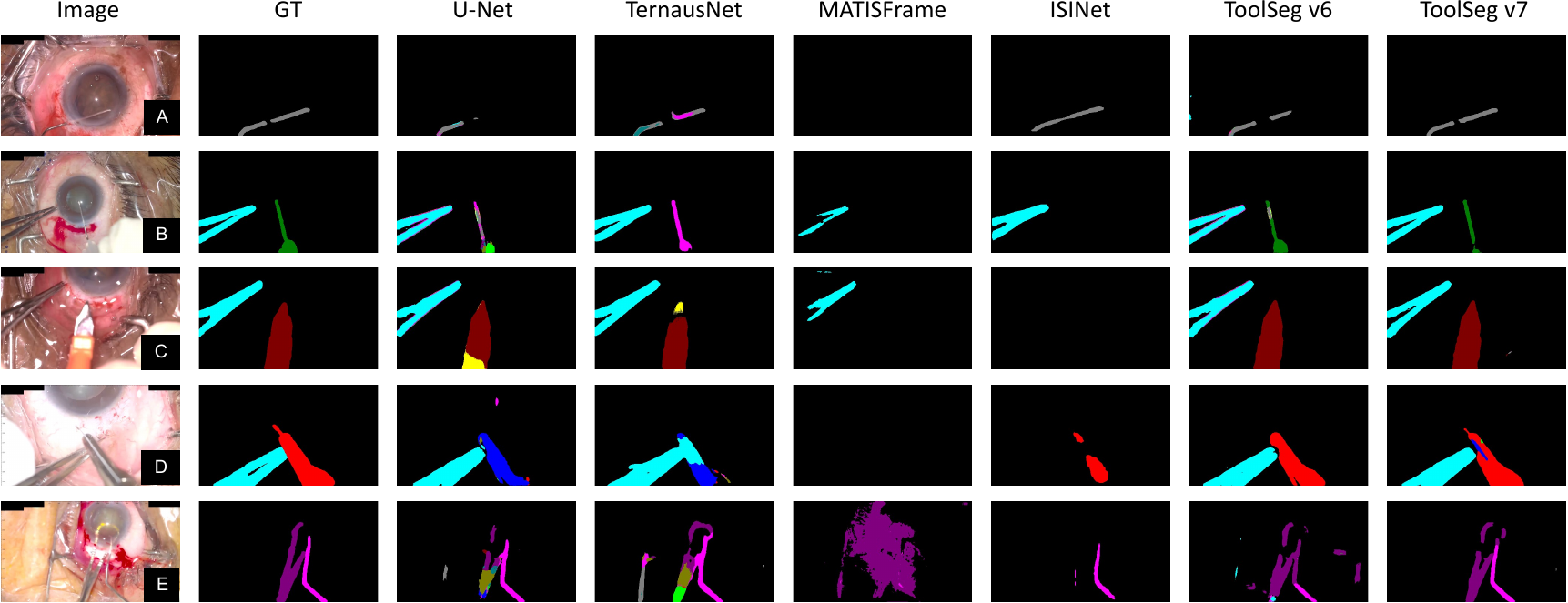}
    \caption{Qualitative results of our proposed \textit{ToolSeg} with SOTA methods. (A) \textcolor[HTML]{808080}{Hydrodissection Cannula}, (B) \textcolor[HTML]{00FFFF}{Hoskins Forceps} and \textcolor[HTML]{008000}{Rhexis Needle}, (C)  \textcolor[HTML]{00FFFF}{Hoskins Forceps} and \textcolor[HTML]{800000}{Keratome}, (D)  \textcolor[HTML]{00FFFF}{Hoskins Forceps} and \textcolor[HTML]{FF0000}{Blade}, and (E) \textcolor[HTML]{800080}{Vectis} and \textcolor[HTML]{FF00FF}{Dialer}.}
    \label{fig:qualitative}
\end{figure*}

\noindent \textbf{Dynamic Feature Blending Factor.} 
To combine spatial and phase-specific information effectively, 
we propose the blending factor $\alpha$, which modulates feature maps \textit{spatially} at the pixel-level based on the input feature map $f$ and phase $p$. 
It is computed as the average of a convolution operation $H$ applied to $f$ and a learnable phase embedding $\eta_p$: 
\[\alpha = \frac{H(f) + \eta_p}{2}\]
Here, $H(f)$ captures spatial features of $f$, while $\eta_p$ encodes phase-specific knowledge.
This design enables the network to adapt to phase-dependent variations in feature characteristics. The blending factor functions similarly to an \textit{attention mechanism} by emphasizing important features, but it extends this concept by jointly modulating both spatial and phase-specific information.\\

\begin{table*}[t!]
\centering
\caption{Impact of different components of \textit{ToolSeg}, our proposed approach, on the \msicsDataset{} dataset.}
\label{tab:method-comparison}
\begin{tabular}{cllccc}
\hline
\textbf{ToolSeg Variant} & \textbf{Phase Conditioning} & \textbf{Phase Source} & \textbf{Pseudo Data} & \textbf{IoU} (m $\pm$ std)           & \textbf{DSC} (m $\pm$ std)    \\
\hline
v0               & -                           & -                     & -                    & 40.90 $\pm$ 3.4                             & 50.66 $\pm$ 3.9                              \\
v1               & -                           & -                     & Yes                  & 48.68 $\pm$ 4.9                             & 58.29 $\pm$ 5.0                              \\
v2               & PCD-Base                        & Predicted MSTCN++     & -                    & 46.58 $\pm$ 5.5                             & 55.40 $\pm$ 5.8                              \\
v3               & PCD-Gated                  & Predicted MSTCN++     & -                    & 48.77 $\pm$ 5.5                             & 57.52 $\pm$ 5.4                              \\
v4               & PCD-Gated                  & Predicted MSTCN++     & Yes                  & 54.32 $\pm$ 4.4                           & 62.70 $\pm$ 4.3                               \\
v5               & PCD-Base                        & Ground Truth          & -                    & 54.26 $\pm$ 4.6                             & 62.98 $\pm$ 4.7                              \\
v6               & PCD-Gated                  & Ground Truth          & -                    & 56.13 $\pm$ 4.5                             & 64.76 $\pm$ 4.5                              \\
v7               & PCD-Gated                  & Ground Truth          & Yes                  & \textbf{61.62 $\pm$ 3.8}                             & \textbf{69.96 $\pm$ 3.8}                              \\ 
\hline
\end{tabular}
\end{table*}

\noindent
\textbf{Context-Aware Adaptive Gating.} Building upon the computed blending factor $\alpha$, we design the CGate mechanism to fuse the phase-modulated features $f^\prime$, which emphasizes phase-relevant details, with the original input feature map $f$. The final output feature map $f^{\prime\prime}$ is computed as:
\[ f^{\prime\prime} = f^\prime \cdot \alpha + f \cdot (1 - \alpha)\]
Here, $\alpha$ acts as a dynamic gate, controlling the relative contribution of phase-modulated and original features. A higher $\alpha$ emphasizes phase-specific information that are distinct for each surgical phase for tool identification, whereas a lower $\alpha$ ensures that original spatial information is preserved. This adaptive gating mechanism enables the network to flexibly combine phase-relevant and spatially important features, enhancing robustness and performance across varying surgical contexts.

\subsection{Semi-Supervised Learning with SAM 2}

Surgical tool segmentation models typically rely on \textit{sparsely} annotated video datasets, where only a small fraction of frames are labeled with ground truth masks.
Since only trained professionals, such as surgeons, possess the domain expertise needed to provide precise, clinically relevant labels, the annotation process is labor-intensive, time-consuming, and costly.
Consequently, a majority of frames remain underutilized. To address this, we propose a simple yet effective pseudo-label-based semi-supervised learning framework, leveraging SAM 2 foundation model and targeted selection of unlabeled data.\\

\noindent \textbf{Prompting-based Pseudo-Label Generation.}
To generate high-quality pseudo-labels, we use the state-of-the-art interactive foundation model, Meta's Segment Anything Model 2 (SAM 2)~\cite{ravi2024sam}.
SAM 2 is a promptable model capable of processing point-based and bounding box based prompts.
Additionally, SAM 2 can propagate the mask from a seed frame across the video length, functioning similar to object tracking. These capabilities can be leveraged to generate efficient and accurate pseudo labels from sparse high quality annotated frames.

For each manually annotated frame, we follow a two-step process for each tool present in that frame:
\begin{enumerate}
    \item Sample two points each from the foreground and background regions of the ground truth mask for the specific tool. Use these points as initial prompts to generate a preliminary prediction mask with SAM 2.
    \item Iteratively refine the predicted mask by following the below steps:
    \begin{enumerate}
        \item Divide it into four regions: true positive, true negative, false positive, and false negative, based on the ground truth mask, and then sample two points from each region.
        \item Use these eight points as prompts, with positive and negative labels to guide SAM 2 towards a more accurate tool prediction mask.
    \end{enumerate}
\end{enumerate}

\noindent
We retain the set of point prompts that generated the tool mask with the highest overlap with the ground truth mask.
This set is then used to generate pseudo-labels for neighboring unlabeled frames. Specifically, we propagate the mask forward and backward across 90 frames, sampling every 30th frame to avoid redundancy. This method leverages the temporal coherence in surgical videos, generating six additional pseudo-labeled frames for each labeled frame.
Using this method, we generated 20,878 pseudo-labeled frames from our base dataset of 3,527 annotated frames.\\

\noindent \textbf{Self-Training using Pseudo Labels.} We use a semi-supervised training strategy, leveraging a combination of annotated and pseudo-labeled data. Pseudo labels provide weak supervision, allowing the model to learn from a larger low-quality dataset. This is followed by fine-tuning on high-quality annotated data for strong supervision, ensuring the model retains alignment with the expert labeled data.
\section{Experiments and Discussion}

\begin{table}[t!]
\centering
\caption{Effect of varying sizes of manually annotated \msicsDataset{} data on pseudo-labeling performance}
\label{tab:pseudo-data-ablation}
\begin{tabular}{rccc}
\hline
\multicolumn{1}{c}{\textbf{\begin{tabular}[c]{@{}c@{}}GT\\ Data\end{tabular}}} &
  \textbf{\begin{tabular}[c]{@{}c@{}}Pseudo\\ Data\end{tabular}} &
  \textbf{IoU} (m $\pm$ std) &
  \textbf{DSC} (m $\pm$ std) \\ \hline
100\% & No  & 56.13 $\pm$ 4.5 & 64.76 $\pm$ 4.5 \\
25\%  & Yes & 51.39 $\pm$ 5.5 & 60.36 $\pm$ 5.3 \\
50\%  & Yes & 57.11 $\pm$ 4.6 & 65.92 $\pm$ 3.9 \\
100\% & Yes & \textbf{61.62 $\pm$ 3.8} & \textbf{69.96 $\pm$ 3.8} \\ \hline
\end{tabular}%
\end{table}

\subsection{Experimental Setup}
\label{sec:experiments}
\noindent \textbf{Data splits.}
We evaluate our model using five-fold cross-validation. The dataset is split into five equal test sets at the video level, with the remaining data randomly divided into train and validation sets at an $80$:$20$ ratio. More details are provided in the supplementary material.\\

\noindent \textbf{Evaluation Metrics.}
We evaluate the segmentation performance of our model using two metrics: Intersection over Union (IoU) and Dice Similarity Coefficient (DSC).\\

\noindent \textbf{Implementation Details.}
Our model is based on a U-Net encoder-decoder architecture with four stages and a bottleneck layer. Training uses the AdamW optimizer with an initial learning rate of $1e-4$. All code is implemented in PyTorch. The experiments run on an NVIDIA A100 GPU with a batch size of 16 for up to 100 epochs, with early stopping applied at a patience of 10. For phase recognition, we use the publicly available MS-TCN++ model, trained on our surgical video dataset. Further implementation details are available in the supplementary material.

\begin{table}[t!]
\centering
\caption{Comparison of \textit{ToolSeg}, our proposed approach, against the state-of-the-art methods on the \msicsDataset{} dataset.}
\label{tab:sota-method-comparison}
\begin{tabular}{lcc}
\hline
\textbf{Method}                                         & \textbf{IoU} (m $\pm$ std) & \textbf{DSC} (m $\pm$ std) \\ \hline
U-Net~\cite{ronneberger2015u}                    & 40.90 $\pm$ 3.4                             & 50.66 $\pm$ 3.9                              \\
TernausNet~\cite{iglovikov2018ternausnet}                                              & 42.76 $\pm$ 5.8                             & 52.03 $\pm$ 6.0                                \\
ISINet~\cite{gonzalez2020isinet}  & 27.55 $\pm$ 3.1                             & 37.60 $\pm$ 3.7                               \\
MATIS Frame~\cite{ayobi2023matis} & 11.41 $\pm$ 6.8                             & 18.24 $\pm$ 10.1                             \\ \hline
ToolSeg v4                                 & \textbf{54.32 $\pm$ 34.4}                            & \textbf{62.70 $\pm$ 4.3}                              \\
ToolSeg v7                                 & \textbf{61.62 $\pm$ 3.8}                             & \textbf{69.96 $\pm$ 3.8}                              \\
\hline
\end{tabular}
\end{table}

\subsection{Results and Analysis}

\noindent \textbf{Evaluating ToolSeg Variants.}
We assess the contributions of each component in our proposed solution by constructing multiple model variants, with results summarized in Table~\ref{tab:method-comparison}. Our baseline UNet-based encoder-decoder model without phase conditioning or pseudo data (v0) achieves a mean IoU of 40.90 and DSC of 50.66.\\\\

\begin{table*}
\centering
\caption{Comparison of various approaches for segmenting different tools in the \msicsDataset{} dataset based on DSC. Background constitutes an average of 94.51\% of Pixel Occupancy per frame.}
\label{tab:classwise-comparison}
\resizebox{\linewidth}{!}{%
\begin{tabular}{lrcccccc}
\hline
\textbf{Tools} & \textbf{\#Instances} & \textbf{Pixel Occ.} & \textbf{ToolSeg v0} & \textbf{ToolSeg v3} & \textbf{ToolSeg v4} & \textbf{ToolSeg v6} & \textbf{ToolSeg v7} \\ \hline
Blade                   & 387  & 1.70 & 46.44 $\pm$ 11.2 & 63.35 $\pm$ 03.8 & 63.42 $\pm$ 08.5 & 69.02 $\pm$ 06.3 & \textbf{74.82 $\pm$ 04.9} \\
Cautery                 & 278  & 1.13 & 63.81 $\pm$ 09.5 & 60.28 $\pm$ 06.8 & 70.90 $\pm$ 10.3 & 73.38 $\pm$ 07.1 & \textbf{75.46 $\pm$ 06.3} \\
Conjunctival Scissors   & 156  & 1.13 & 61.89 $\pm$ 08.4 & 68.12 $\pm$ 04.6 & 75.58 $\pm$ 05.9 & 76.00 $\pm$ 06.4 & \textbf{82.29 $\pm$ 02.7} \\
Crescent Blade          & 390  & 1.62 & 67.38 $\pm$ 22.2 & 67.41 $\pm$ 16.3 & 73.66 $\pm$ 17.1 & 76.14 $\pm$ 15.8 & \textbf{79.98 $\pm$ 13.9} \\
Dialer                  & 343  & 0.75 & 40.33 $\pm$ 12.5 & 41.54 $\pm$ 07.8 & 49.67 $\pm$ 10.2 & 41.23 $\pm$ 08.5 & \textbf{58.96 $\pm$ 08.2} \\
Hoskins Forceps         & 2005 & 3.44 & 78.34 $\pm$ 07.3 & 78.73 $\pm$ 05.2 & 82.15 $\pm$ 05.4 & 80.62 $\pm$ 04.3 & \textbf{84.34 $\pm$ 05.2} \\
Hydrodissection Cannula & 339  & 0.73 & 36.50 $\pm$ 15.9 & 36.70 $\pm$ 08.1 & 49.55 $\pm$ 08.8 & \textbf{57.57 $\pm$ 04.0} & 56.18 $\pm$ 07.5 \\
Keratome                & 231  & 1.81 & 46.24 $\pm$ 14.8 & 67.92 $\pm$ 15.5 & 70.17 $\pm$ 11.7 & 75.71 $\pm$ 05.4 & \textbf{77.84 $\pm$ 04.7} \\
Rhexis Needle           & 86   & 0.71 & 13.98 $\pm$ 07.0 & 30.27 $\pm$ 11.5 & 27.56 $\pm$ 07.0 & 38.70 $\pm$ 11.5 & \textbf{47.22 $\pm$ 14.0} \\
Sideport                & 240  & 2.69 & 68.69 $\pm$ 11.0 & 63.73 $\pm$ 11.9 & 74.93 $\pm$ 10.9 & 76.79 $\pm$ 08.0 & \textbf{77.57 $\pm$ 09.8} \\
Simcoe Cannula          & 319  & 1.70 & 46.90 $\pm$ 15.8 & 57.30 $\pm$ 11.5 & 66.11 $\pm$ 08.9 & 65.12 $\pm$ 13.4 & \textbf{72.03 $\pm$ 06.5} \\
Vectis                  & 152  & 1.96 & 41.67 $\pm$ 13.9 & 44.72 $\pm$ 15.9 & 56.17 $\pm$ 14.5 & 55.36 $\pm$ 12.7 & \textbf{61.80 $\pm$ 13.1} \\
Visco Cannula           & 396  & 1.59 & 46.42 $\pm$ 06.9 & 40.08 $\pm$ 04.8 & 55.18 $\pm$ 06.0 & 56.28 $\pm$ 03.2 & \textbf{61.04 $\pm$ 04.6} \\ \hline
\end{tabular}%
}
\end{table*}

\noindent \textit{\textbf{(i) Impact of Phase Conditioning.}}
To evaluate the impact of phase information, we first test our model with ground truth (GT) phases as a prior, establishing an upper-bound performance. Incorporating GT phase (either PCD-Basic in v5 or PCD-Gated in v6) significantly increases segmentation accuracy, with IoU gains of $32.67\%$ to $37.24\%$ and DSC gains of $24.32\%$ to $27.83\%$ over the baseline. Since ground truth phase information may not be available in practical applications, we also test conditioning on predicted phases from MS-TCN++ (v2 and v3), which still improve IoU by $13.89\%$ to $19.24\%$ and DSC by $9.36\%$ to $13.54\%$. These results demonstrate that the model can effectively leverage phase information, whether from ground truth or predictions.\\

\noindent \textit{\textbf{(ii) Effect of Pseudo-labeled Data.}}
Adding pseudo-labeled data alone (v1) improves IoU by $19.02\%$ and DSC by $15.06\%$, showing the benefit of utilizing otherwise unlabeled video frames. Combining PCD-Gated model with pseudo-labeled data (v4 and v7) yields substantial gains, increasing IoU by $32.81\%$ to $50.66\%$ and DSC by $23.77\%$ to $38.1\%$. This highlights the complimentary strengths of phase information and semi-supervised learning.\\

\noindent \textit{\textbf{(iii) Comparison of Conditioning Mechanisms.}}
We compare PCD-Basic and PCD-Gated, and found that gating consistently outperforms basic conditioning by $3.5\%$ in IoU and $2.8\%$ in DSC with GT phases, and
by $4.7\%$ in IoU and $3.8\%$ in DSC when using predicted phases. These results highlight gating's ability to adaptively control information flow is particularly beneficial when using predicted phases, as it enables the model to filter out irrelevant signals, thus enhancing robustness and segmentation accuracy.\\

\noindent \textbf{Pseudo-labeling with Lesser Manually Annotated Data.}
We evaluate the impact of our semi-supervised setup ---which involves a prompting-based pseudo-labeling method---by training the model with varying proportions of manually annotated data: $25\%$, $50\%$ and $100\%$. Results in Table~\ref{tab:pseudo-data-ablation} show that using only $50\%$ of the manually labeled data combined with pseudo-labeled data outperforms a fully supervised model trained with $100\%$ labeled data. Moreover, with just $25\%$ labeled data, the performance remains comparable to the fully supervised model. These findings highlight the effectiveness of our approach in achieving strong performance with significantly reduced annotation requirements.\\

\noindent \textbf{SOTA Benchmarking Results.}
We benchmark our proposed \textit{ToolSeg} model against several state-of-the-art models,
including U-Net~\cite{ronneberger2015u}, TernausNet~\cite{iglovikov2018ternausnet}, ISINet~\cite{gonzalez2020isinet} and MATIS-Frame~\cite{ayobi2023matis}---all of which have demonstrated effectiveness in various surgical contexts. The results, summarized in Table~\ref{tab:sota-method-comparison}, show that \textit{ToolSeg} with gated phase conditioning and semi-supervised learning achieves significantly higher performance compared to all benchmarked models. Among the baselines, TernausNet achieves the highest performance with an IoU of $42.76$ and DSC of $52.03$, followed by U-Net with an IoU of $40.90$ and DSC of $52.03$. ISINet and MATIS-Frame, which has been found to excel in gasterointestinal surgery datasets, exhibit significantly lower IoU scores of $27.55$ and $11.41$, respectively. This discrepancy highlights the unique challenges of ocular surgery, where tools are smaller, often resemble each other, and blend into complex anatomical backgrounds, making segmentation more challenging.\\

\noindent \textbf{Tool Specific Performance Trends.}
The performance of our approach on individual surgical tools is presented in Table~\ref{tab:classwise-comparison}.
Tools like the Blade, Keratome, and Rhexis Needle show significant improvements with phase priors (ToolSeg v3 and v6 versus v0), highlighting that phase information is especially beneficial for tools with distinct roles or appearances in specific surgical phases (Figure~\ref{fig:qualitative}). In contrast, tools such as the Hoskins Forceps and Dialer show smaller improvements, as they are used across multiple surgical phases.
Semi-supervised learning further boosts performance, especially for underrepresented tools. 
For instance, the Rhexis Needle and Dialer show substantial gains of $21.99\%$ and $43.01\%$ (ToolSeg v6 versus v7), likely due to their lower instance counts and pixel occupancy in the base labeled dataset, thus benefiting from additional unlabeled frames~\ref{fig:qualitative}. 
Tools with higher pixel occupancy, such as the Hoskins Forceps, show smaller relative improvements, possibly due to their relatively significant presence in the base dataset. 
These findings suggest that phase priors are most beneficial for phase-dependent tools, while semi-supervised learning helps improve the segmentation of tools with fewer instances or lower pixel presence.\\

\begin{table}
\centering
\caption{Comparison of \textit{ToolSeg}, our proposed approach, against the state-of-the-art methods on the CaDIS dataset.}
\label{tab:cadis-results}
\begin{tabular}{lcc}
\hline
\textbf{Model}                            & \textbf{IoU} (m) & \textbf{DSC} (m) \\ \hline
U-Net~\cite{ronneberger2015u}             & 52.69            & 62.84            \\
TernausNet~\cite{iglovikov2018ternausnet} & 46.47            & 55.22            \\
ISINet~\cite{gonzalez2020isinet}          & 11.51            & 15.41            \\
MATISFrame~\cite{ayobi2023matis}          & 25.59            & 34.43            \\ \hline
ToolSeg v1                                & 54.65            & 64.36            \\
ToolSeg v6                                & \textbf{60.73}            & \textbf{68.63}            \\ 
ToolSeg v7                                & 59.05            & 67.72            \\ \hline
\end{tabular}
\end{table}
\noindent \textbf{Generalization to CaDIS Dataset.} To assess the generalizability of our proposed method, we evaluate it on the CaDIS dataset~\cite{grammatikopoulou2021cadis}, which focuses on Phaco cataract surgeries. We consider 13 tools across 18 surgical phases. Using the ToolSeg variant with the best overall performance (v6), our model achieves a mean IoU of $60.73\%$ and a DSC of $68.63\%$, significantly outperforming SOTA models U-Net, TernausNet and MATIS-Frame by $9.21\%$, $24.28\%$ and $99.33\%$ in DSC, respectively, as shown in Table~\ref{tab:cadis-results}. The semi-supervised setup (v1) alone improves the baseline (v0) by $2.42\%$ in DSC, and incorporating gated phase conditioning with GT phases (v6) provides a substantial $15.26\%$ IoU boost over the baseline. These results demonstrate that our approach generalizes effectively to other surgical datasets, achieving notable performance gains.\\

\subsection{Limitations and Future Works.}
While the current work advances surgical tool segmentation, several limitations exist that open up opportunities for future research.
First, the current dataset is limited in annotated video volume and diversity; expanding this to include more complicated and anomalous surgeries would improve the model's generalizability. 
Second, although quality control and denoising steps were included in the annotation process, some noise may still be present due to the use of a SAM-based annotation tool.
Third, a performance gap exists between predictions based on estimated phases versus ground truth phases. This gap could be addressed by employing advanced phase recognition algorithms, which is beyond the scope of this paper. End-to-end training that jointly optimize phase recognition and tool segmentation also warrants exploration.
Fourth, our pseudo-labeling approach, while resulted in performance boost, is currently limited by sampling every 30th frame, which may overlook relevant frames for generating high-quality pseudo-labels. Refining this with a dynamic frame selection algorithm could further improve performance.

Beyond tool classification and phase recognition, these tasks are only parts of the larger objective of analyzing surgical procedures in real-time for both training purposes and anomaly detection.
Future user-centric research should explore how best to utilize phase and segmented tool masks for providing offline actionable training and online feedback to surgeons. This would involve exploring how surgeons would interact with and interpret this information to optimize their performance.
Regarding anomaly detection, our current method lays the groundwork for identifying potential surgical errors, such as tool misuse, incorrect phase sequencing, or deviations from standard surgical practices (e.g., larger-than-usual incisions).
Real-time alerts for such deviations have the potential to significantly enhance surgical safety.
\section{Conclusion}
In this paper, we present \msicsDataset{}, the first comprehensive dataset on Manual Small-Incision Cataract Surgery, addressing a critical gap in AI-driven surgical video analysis in a widely performed but underexplored surgical procedure. The dataset comprises of 3,527 frames from 53 surgical videos with surgical phase and tool annotations.
Our benchmarking revealed the limitations of existing tool segmentation models in accurately classifying and segmenting MSICS tools.
To address this, we introduce \textit{ToolSeg}, a novel phase-informed tool segmentation framework, which demonstrates significant performance improvements by leveraging surgical phase information as a prior. Additionally, we employ a SAM-2-based label propagation strategy to expand the dataset to 24,405 frames, underscoring the scalability of automated annotation techniques to reduce manual labeling efforts. Our experiments further validate the generalizability of ToolSeg and the pseudo-labeling approach to other cataract surgery settings.
We hope these results establish a solid foundation for future work in surgical tool segmentation, with the potential to enhance automated analysis in MSICS and similar procedures.
{
    \small
    \bibliographystyle{ieeenat_fullname}
    \bibliography{main}
}
\clearpage

\setcounter{page}{1}
\setcounter{section}{0}
\maketitlesupplementary

\section{Background}

\begin{figure}[b!]
    \centering
    \includegraphics[width=1.0\linewidth]{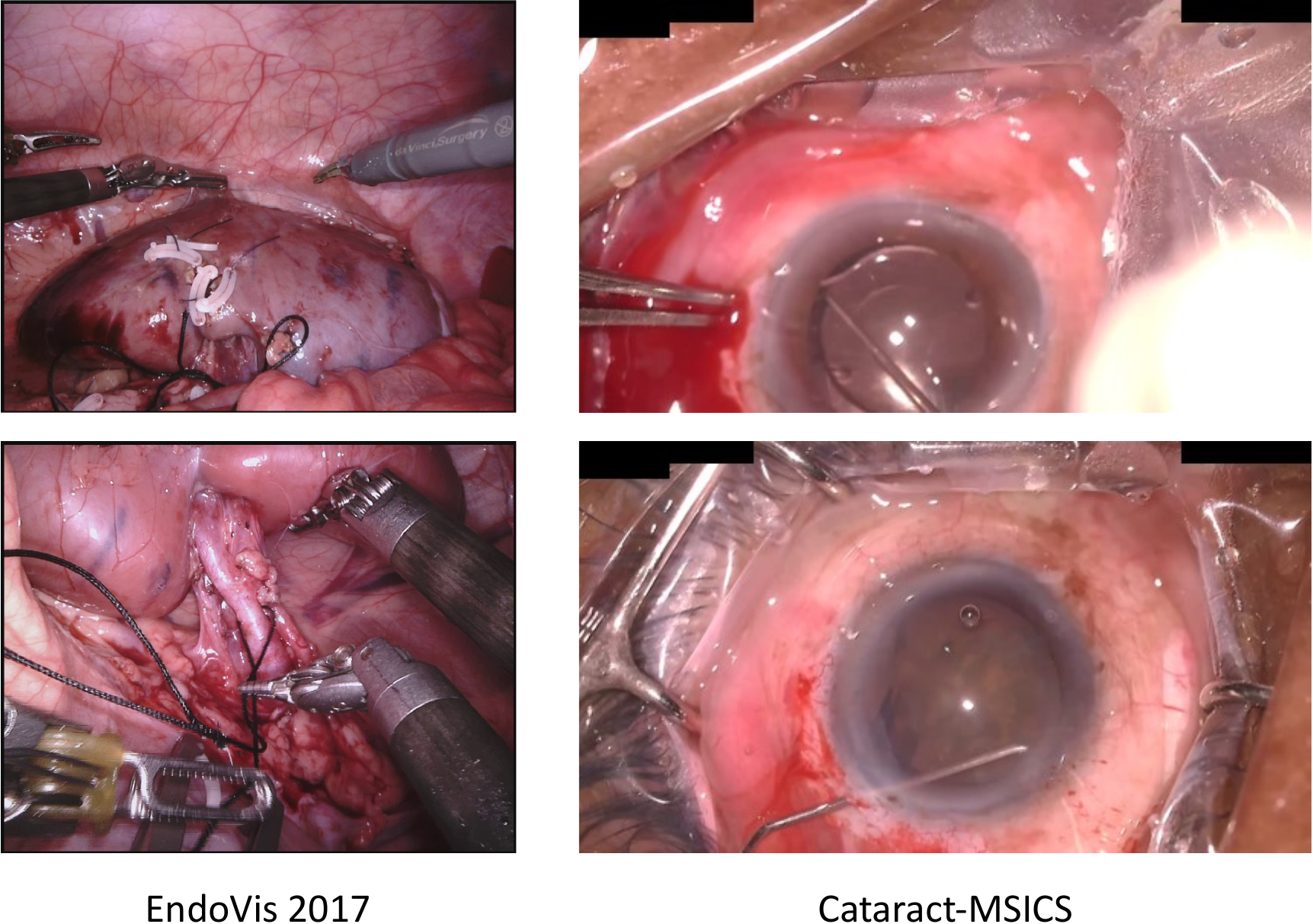}
    \caption{Comparison of surgical scenes: a laparoscopic procedure (EndoVis 2017) versus an ocular procedure (\msicsDataset{}). The \msicsDataset{} example showcases challenges including high surface reflectivity and minute tool size, while the EndoVis image illustrates the relatively simpler environment with larger instruments and a clear background.}
    \label{fig:endovis_msics}
\end{figure}

\subsection{Ocular vs Laparoscopic Surgery}

Most computer vision datasets in surgery focus on laparoscopic procedures, particularly in gastroenterology, leaving other domains like ocular surgeries comparatively underrepresented.

Ocular surgeries pose unique challenges due to the highly confined surgical field, where microscale instruments operate within a restricted maneuvering space. That increases the complexity in precisely identifying and tracking instruments. The reflective corneal and scleral surfaces introduce glare, obscuring both instruments and anatomical details. Dynamic factors like eye movements and fluid manipulation make the process even more challenging.
In contrast, laparoscopic surgeries offer a relatively less constrained environment, with larger instruments occupying a significant portion of the visual field, simplifying detection and tracking tasks. The diffuse illumination and soft tissue environment typical of laparoscopic procedures also reduce visual obstructions like glare. 
Figure~\ref{fig:endovis_msics} illustrates these differences using examples from the \msicsDataset{} and EndoVis 2017 datasets,
highlighting the unique complexities of ocular surgeries, underscoring the need for advanced computer vision solutions tailored to this domain.

\subsection{MSICS Details}

\begin{table*}[]
\centering
\caption{Surgical workflow of a routine MSICS procedure. Note: Tools with fewer than 50 instances, including MC Forceps, Needle, Rhexis Forceps, Trypan blue dye, and Needle Holder, were excluded from the dataset.}
\label{tab:msics_details}
\begin{tabular}{c|c|l|l}
\hline
\textbf{Stages} &
  \textbf{Phases} &
  \textbf{Description} &
  \textbf{Tools used} \\ \hline
\multirow{4}{*}{\begin{tabular}[c]{@{}c@{}}Tunnel\\ construction\end{tabular}} &
  Peritomy &
  \begin{tabular}[c]{@{}l@{}}Forceps and scissors are used to make the\\ first incision in the eye.\end{tabular} &
  \begin{tabular}[c]{@{}l@{}}Hoskins Forceps\\ MC Forceps\\ Conjunctival Scissors\end{tabular} \\ \cline{2-4} 
 &
  Scleral Groove &
  \begin{tabular}[c]{@{}l@{}}Groove helps provide a stable base for the\\ incision and tunnel construction.\end{tabular} &
  \begin{tabular}[c]{@{}l@{}}Blade\\ Crescent Blade\\ Hoskins Forceps\end{tabular} \\ \cline{2-4} 
 &
  Incision &
  Initial partial thickness cut is made on the sclera. &
  \begin{tabular}[c]{@{}l@{}}Blade\\ Crescent Blade\\ Hoskins Forceps\end{tabular} \\ \cline{2-4} 
 &
  Tunnel &
  \begin{tabular}[c]{@{}l@{}}Sclerocorneal tunnel is made with the help of a\\ crescent blade.\end{tabular} &
  Crescent Blade \\ \hline
\multirow{4}{*}{\begin{tabular}[c]{@{}c@{}}Capsulorhexis \\ (Anterior capsule \\ opening)\end{tabular}} &
  Sideport &
  Side incision is made using sideport knife. &
  \begin{tabular}[c]{@{}l@{}}Sideport\\ Hoskins Forceps\end{tabular} \\ \cline{2-4} 
 &
  AB Injection \& Wash &
  \begin{tabular}[c]{@{}l@{}}An air bubble is inserted along with blue dye\\ for staining, followed by a wash.\end{tabular} &
  \begin{tabular}[c]{@{}l@{}}Hydro. Cannula\\ Trypan blue dye\end{tabular} \\ \cline{2-4} 
 &
  OVD Injection &
  Viscoelastic liquid is injected. &
  \begin{tabular}[c]{@{}l@{}}Visco Cannula\\ Needle (rarely)\end{tabular} \\ \cline{2-4} 
 &
  Capsulorhexis &
  Opening of the anterior capsule. &
  \begin{tabular}[c]{@{}l@{}}Rhexis Needle\\ Rhexis Forceps\end{tabular} \\ \hline
Tunnel Entry &
  Main Incision Entry &
  The main tunnel entry is done in this phase. &
  \begin{tabular}[c]{@{}l@{}}Keratome\\ Blade\end{tabular} \\ \hline
\multirow{4}{*}{\begin{tabular}[c]{@{}c@{}}Lens break and \\ removal\end{tabular}} &
  Hydroprocedure &
  \begin{tabular}[c]{@{}l@{}}The cataractous nucleus is broken from its\\ surrounding capsular bag using a fluid wave\\ delivered with a jerk, followed by viscoelastic\\ liquid injection.\end{tabular} &
  Hydro. Cannula \\ \cline{2-4} 
 &
  Nucleus Prolapse &
  \begin{tabular}[c]{@{}l@{}}The cataractous nucleus is displaced from\\ the capsular bag into the anterior chamber.\end{tabular} &
  \begin{tabular}[c]{@{}l@{}}Dialer\\ Visco Cannula\end{tabular} \\ \cline{2-4} 
 &
  Nucleus Delivery &
  \begin{tabular}[c]{@{}l@{}}The cataractous lens is removed from the eye\\ using vectis and dialer.\end{tabular} &
  \begin{tabular}[c]{@{}l@{}}Vectis\\ Dialer\end{tabular} \\ \cline{2-4} 
 &
  Cortical Wash &
  \begin{tabular}[c]{@{}l@{}}The cortex is washed to get rid of any residue\\ and viscoelastic liquid is inserted.\end{tabular} &
  Simcoe Cannula \\ \hline
\multirow{2}{*}{IOL lens insertion} &
  OVD, IOL Insertion &
  \begin{tabular}[c]{@{}l@{}}New IOL lens is inserted in the eye, using\\ forceps and dialer to position it.\end{tabular} &
  \begin{tabular}[c]{@{}l@{}}Dialer\\ Hoskins Forceps\end{tabular} \\ \cline{2-4} 
 &
  OVD Wash &
  \begin{tabular}[c]{@{}l@{}}Excess viscoelastic material in the eye is\\ washed off.\end{tabular} &
  Simcoe Cannula \\ \hline
\multirow{3}{*}{Wound closure} &
  Stromal Hydration &
  \begin{tabular}[c]{@{}l@{}}This phase is used for forming and hydrating\\ anterior chamber.\end{tabular} &
  Hydro. Cannula \\ \cline{2-4} 
 &
  Tunnel Suture &
  The tunnel is closed using suture material. &
  \begin{tabular}[c]{@{}l@{}}Needle Holder\\ Hoskins Forceps\\ Suture Material\end{tabular} \\ \cline{2-4} 
 &
  Conjunctival Cautery &
  \begin{tabular}[c]{@{}l@{}}The wound is cauterized using hoskins forceps\\ and cautery.\end{tabular} &
  \begin{tabular}[c]{@{}l@{}}Cautery\\ Hoskins Forceps\end{tabular} \\ \hline
\end{tabular}
\end{table*}
The surgical workflow of a standard MSICS procedure is outlined in Table~\ref{tab:msics_details}, detailing the stages, phases, and tools involved in each step. Key stages include tunnel construction, capsulorhexis, lens removal, IOL insertion, and wound closure, with each stage comprising of multiple surgical phases.
The phases specify the tools used and their functions.
These structured tool-phase patterns highlight the importance of surgical phase information as a prior in tool segmentation models.

\section{Dataset}

\begin{figure}[t!]
    \centering
    \fbox{\includegraphics[width=1.0\linewidth]{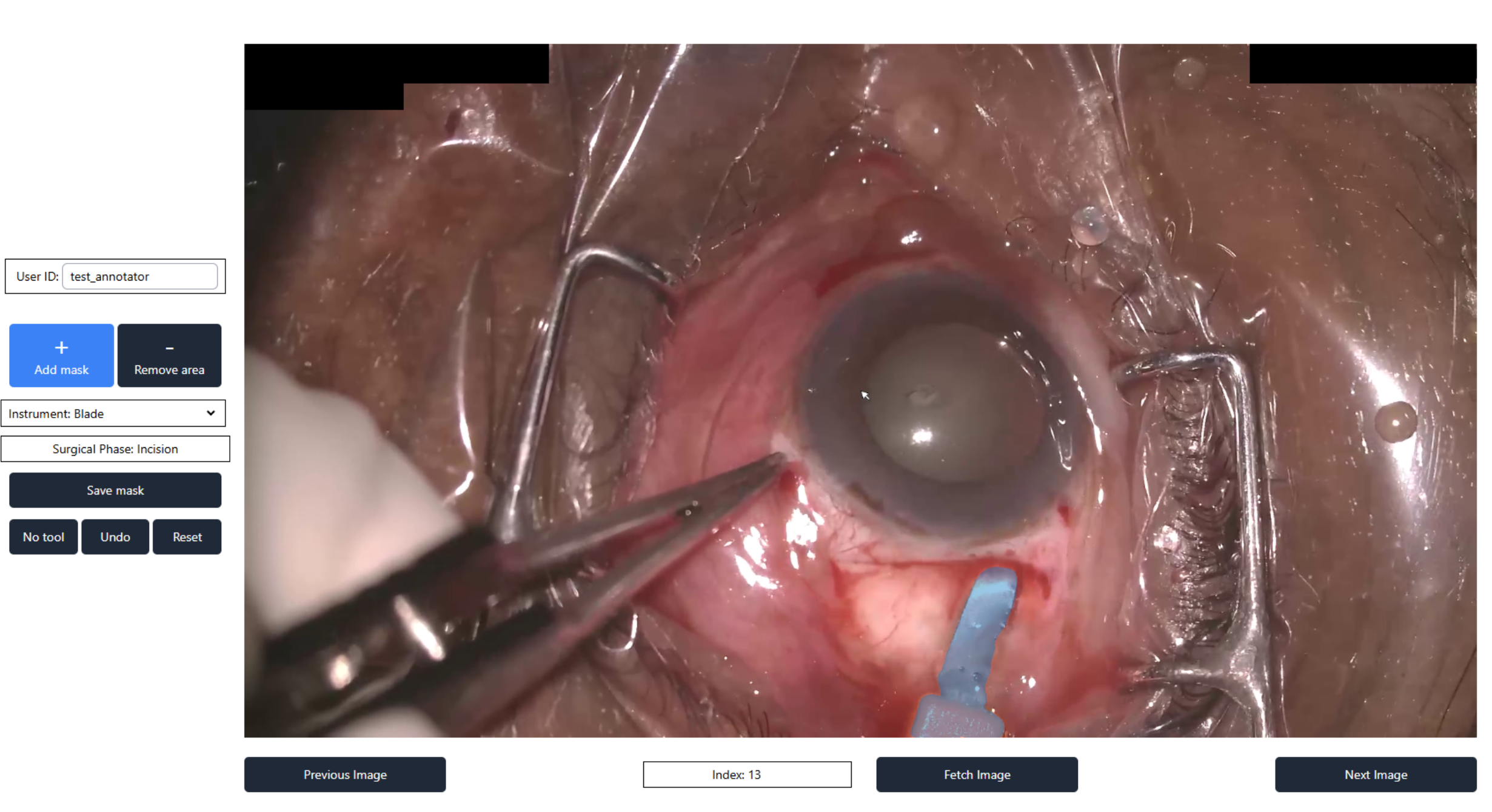}}
    \caption{SAM based annotation web tool.}
    \label{fig:ann_tool}
\end{figure}

\begin{figure*}[!t]
    \centering
    \includegraphics[width=1.0\linewidth]{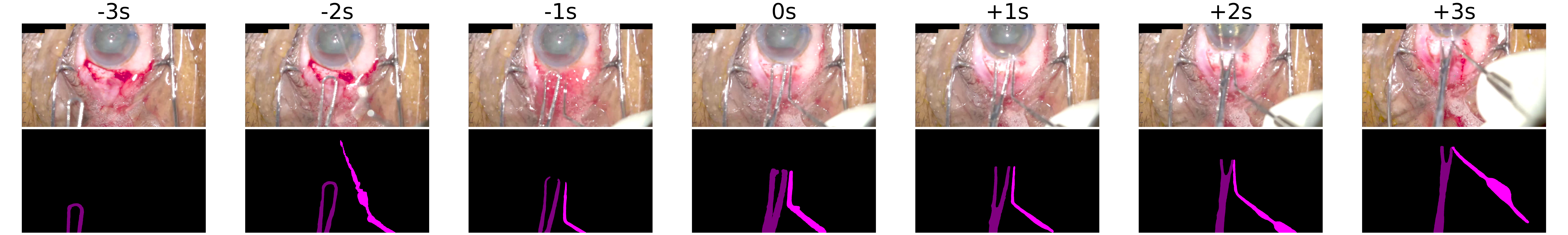}
    \includegraphics[width=1.0\linewidth]{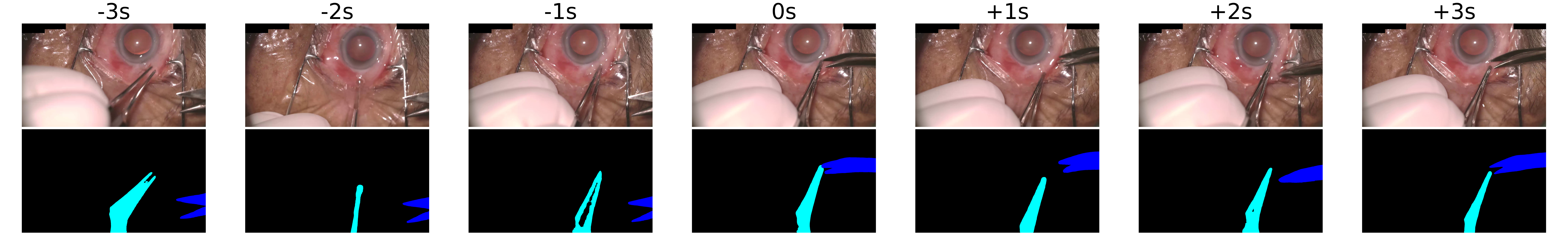}
    \caption{Surgical tool label propagation using SAM 2}
    \label{fig:sam2_propagation}
\end{figure*}

\subsection{Annotation Web Tool and Data Processing}
The annotation tool, built using React.js and Node.js and leveraging Meta's Segment Anything model (SAM)~\cite{kirillov2023segment}, was designed for labeling cataract surgery instruments. Annotators created masks using positive and negative point prompts, and assigned classes via a dropdown menu. To facilitate efficient labeling, the tool provided Reset and Undo functions for easy correction.
Annotations covered 20 surgical tools, with a miscellaneous category for ambiguous cases. Tools with fewer than 50 instances, including the miscellaneous category, were excluded from the training set. Additionally, the tool \textit{speculum}--used to hold the eyelids open--was excluded from annotation as it was consistently visible in every frame and added minimum value.

Annotated masks underwent a denoising process comprising:
\begin{itemize} 
    \item Morphological operations to remove minor artifacts.
    \item Exclusion of connected components smaller than 100 pixels or 10\% of the annotated mask area.
    \item Gaussian blurring (kernel size: 25x25) to smooth masks and reduce noise.
\end{itemize}
Post-denoising, quality control was performed which involved overlap checks and manual comparison between annotators for each batch.
A batch-specific threshold was set to ensure consistency and accuracy. 
All processing steps, including denoising, were performed on the original 1920 x 1080 resolution images to maintain annotation quality.

\section{Implementation Details}
For all our experiments, the model achieving the highest DSC on the validation set is selected for evaluation on the test set. All results are averaged across five folds.

\subsection{Surgical Phase Prediction}
\begin{figure*}[]
    \centering
    \includegraphics[width=0.95\linewidth]{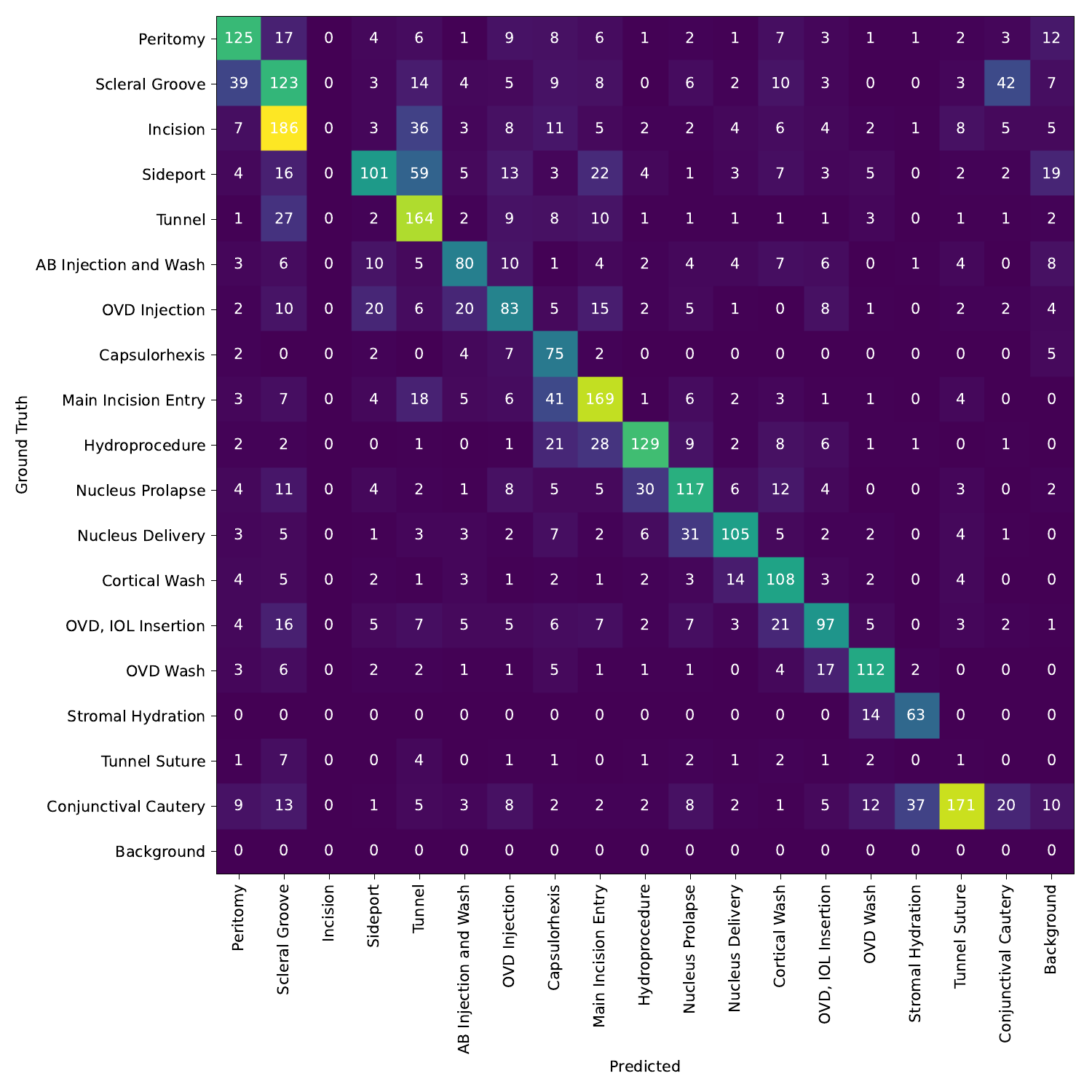}
    \caption{Confusion matrix illustrating the performance of surgical phase prediction on the \msicsDataset{} dataset using MSTCN++ model.}
    \label{fig:conf_matrix}
\end{figure*}

For surgical phase prediction, we use the MSTCN++~\cite{li2020ms} architecture, a temporal action segmentation model. It is a two-stage model leveraging dilated temporal convolutional layers to capture long-range temporal dependencies. The model is trained on I3D features extracted at 15 fps.
To prevent data leakage, MSTCN++ was trained on an internal dataset of 90 surgical videos, independent of the proposed dataset but acquired during the same timeframe. The model was tested on the \msicsDataset{} dataset, and its performance is presented in the confusion matrix shown in Figure~\ref{fig:conf_matrix}. While achieving state-of-the-art surgical phase recognition accuracy is not the goal of this paper, we demonstrate the potential of leveraging an accurate surgical phase recognition model by utilizing ground truth phases. The MSTCN++ model achieves an accuracy of $50.10\%$, precision of $47.60\%$, and an F1 score of $50.23\%$.
The predicted surgical phases are incorporated into our proposed \textit{ToolSeg} framework.
Future advancements in surgical phase recognition would further enhance the performance of \textit{ToolSeg}.

\subsection{Surgical Tool Segmentation Dataset Splits}
From the 53 MSICS videos in our dataset, we generated five stratified data splits, each containing 7-9 test videos. On average, each split consists of 2476 ± 57 training frames, 510 ± 46 validation frames, and 541 ± 31 test frames. Details are available in Table~\ref{table:fold_summary}.
\begin{table}[!]
\centering
\caption{Number of frames in dataset splits for \msicsDataset{}. Note: Number of videos indicated in parentheses.}
\label{table:fold_summary}
\begin{tabular}{c|ccc}
\hline
\textbf{Fold} & \textbf{Train} & \textbf{Validation} & \textbf{Test} \\
\hline
0 & 2387 (37) & 585 (7) & 555 (9) \\
1 & 2434 (38) & 504 (7) & 589 (8) \\
2 & 2542 (39) & 441 (7) & 544 (7) \\
3 & 2506 (39) & 509 (7) & 512 (7) \\
4 & 2513 (39) & 510 (7) & 504 (7) \\
\hline
\end{tabular}
\end{table}

\subsection{SAM 2 based Pseudo-Label Generation}
We used SAM 2-based pseudo-label generation to
augment the labeled dataset by propagating ground truth masks to unlabeled frames for our semi-supervised setup. To assess its effectiveness, we compared 654 common frames in both the expert annotated and pseudo-labeled datasets, derived from 45 videos. The results showed an average DSC of 66.06\% and an IoU of 55.79\%,
indicating effective mask propagation, providing valuable additional supervision for training.
Figure~\ref{fig:sam2_propagation} visualizes this process. For instance, in the first row, the mask propagated to the frame at -2s contains errors due to reflections from the surgeon lamp.

\section{Results}

\subsection{Feature Transformation in PCD-Gated Layer}

\begin{figure*}[b!]
    \centering
    \includegraphics[width=1.0\linewidth]{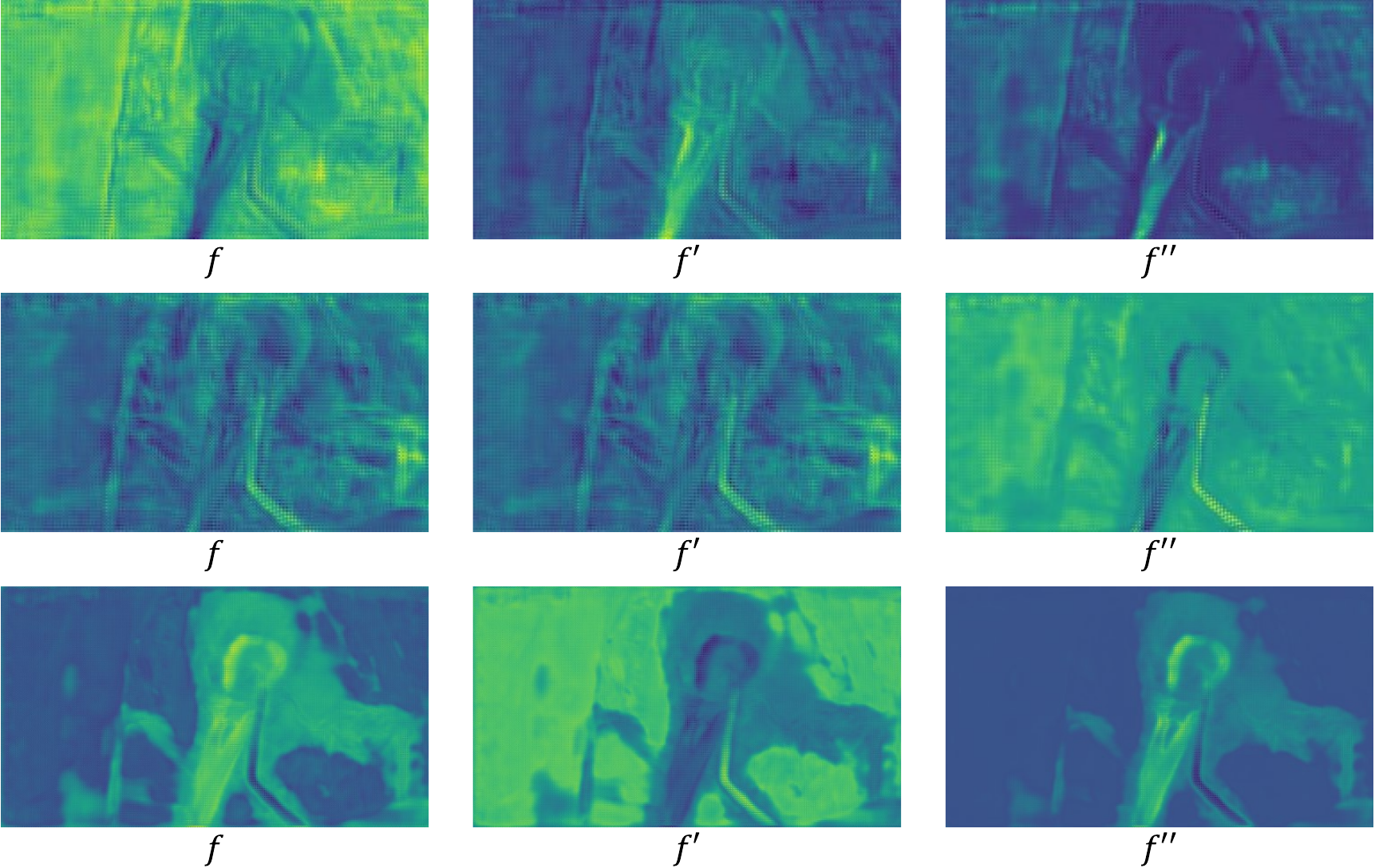}
    \caption{Visualization of feature transformation in the proposed PCD-Gated layer.}
    \label{fig:feat_vis}
\end{figure*}

In this section, we provide visualizations demonstrating how the PCD-Gated layer interacts with features in the latent space.  

Figure~\ref{fig:feat_vis} shows feature maps from the third and the fourth decoder stage, illustrating transformations from \( f \to f' \to f'' \). The example is from the \textit{Nucleus Delivery} phase and includes the \textit{Vectis} and \textit{Dialer}, as outlined in Table~\ref{tab:msics_details}.We observe that the PAFT layer effectively localizes tools by using phase-specific information and their typical shapes and locations. The final output \( f'' \) significantly filters out background noise, enhancing tool visibility. In the first row, the \textit{Vectis} is clearly localized, while in the second row, the \textit{Dialer} is accurately detected. In the third row, \textit{Vectis} is first located and then highlighted by removing the background noise.

\subsection{Additional Results on CaDIS Dataset}
Table~\ref{tab:cadis_segmentation_percentages} summarizes the performance (DSC scores) of various methods on individual tools in the CaDIS dataset. The frequency column indicates the number of occurrences of each tool, ranging from the most frequent \textit{Eye Retractors} (3513 instances), to the least frequent \textit{Cap Forceps} (119 instances).
Our proposed methods, ToolSeg v6 and v7 consistently outperforms other methods across tool classes. ToolSeg v7 achieves the highest scores for \textit{Cap Forceps} (87.92), \textit{Phaco Handpiece} (80.38), and \textit{Primary Knife} (77.87)---tools that span mid- to low-frequency ranges.
Notably, the model achieves significant improvements in challenging classes such as \textit{Hydro Cannula} (65.32) and \textit{Cap Cystotome} (60.42), where baseline models like U-Net and TernausNet struggle.

While high-frequency tools like \textit{Eye Retractors} yield competitive scores for both U-Net (74.49) and ToolSeg v6 (78.43), v7 (72.88), the latter's marked improvements for low-frequency classes, such as \textit{Cap Forceps} (87.92), highlight its robustness.
This indicates our model's ability to generalize effectively across diverse class distributions.

\begin{table*}[t!]
\centering
\caption{Comparison of various approaches for segmenting different tools in the CaDIS dataset based on DSC.}
\label{tab:cadis_segmentation_percentages}
\begin{tabular}{l|c|cccccccc}
\hline
\textbf{Class}   & \textbf{Frequency}                    & \textbf{U-Net} & \textbf{TernausNet} & \textbf{ISINet} & \textbf{MATISFrame} & \textbf{ToolSeg v1} & \textbf{ToolSeg v6} & \textbf{ToolSeg v7} \\ \hline
Eye Retractors      &    3513
              & 74.49          & 77.86               & 15.13           & 2.42                & 77.57              & \textbf{78.43}     & 72.88              \\
Hydro Cannula      &   454
                   & 59.41          & 52.41               & 26.56           & 43.32               & 57.44              & 64.89              & \textbf{65.32}     \\
Visco Cannula     &    604
                   & 52.68          & 36.12               & 18.87           & 35.29               & 56.11              & \textbf{64.42}     & 54.25              \\
Cap Cystotome     &    445
                   & 53.10          & 43.22               & 13.27           & 55.27               & 51.28              & \textbf{65.93}     & 60.42              \\
Rycroft Cannula     &  436
                   & 48.78          & 0.00                & 1.62            & 40.49               & 53.20              & 55.05              & \textbf{55.85}     \\
Bonn Forceps    &      381
                   & 60.96          & 52.56               & 4.00            & 29.01               & 57.96              & 54.53              & \textbf{59.81}     \\
Primary Knife     &   314
                    & 65.79          & 68.15               & 5.96            & 15.72               & 66.99              & 74.26              & \textbf{77.87}     \\
Phaco Handpiece    &      459
                & 70.33          & 70.38               & 29.12           & 40.10               & 76.70              & 79.36              & \textbf{80.38}     \\
Lens Injector    &    783
                    & 75.10          & 75.87               & 51.77           & 45.58               & 75.64              & \textbf{75.85}     & 75.64              \\
A/I Handpiece     &    408
                   & 61.07          & 57.11               & 9.03            & 29.50               & 65.18              & \textbf{71.80}     & 70.84              \\
Secondary Knife   &     303
                 & 53.80          & 68.39               & 0.00            & 13.18               & 53.79              & 58.78              & \textbf{52.67}     \\
Micromanipulator      &       629
            & 59.06          & 59.01               & 0.00            & 27.01               & 67.08              & 62.25              & \textbf{66.58}     \\
Cap Forceps      &     119
                   & 82.44          & 56.89               & 25.09           & 70.72               & 77.82              & 86.68              & \textbf{87.92}     \\ \hline
\end{tabular}
\end{table*}


\end{document}